\documentclass[conference]{IEEEtran}
\IEEEoverridecommandlockouts
\usepackage{cite}
\usepackage{amsmath,amssymb,amsfonts}
\usepackage{algorithmic}
\usepackage{graphicx}
\usepackage{textcomp}
\usepackage{xcolor}
\def\BibTeX{{\rm B\kern-.05em{\sc i\kern-.025em b}\kern-.08em
    T\kern-.1667em\lower.7ex\hbox{E}\kern-.125emX}}
\begin{document}

\title{Deep Learning Approaches for Detecting Adversarial Cyberbullying and Hate Speech in Social Networks\\
{\footnotesize \textsuperscript{}}
}

\author{\IEEEauthorblockN{Sylvia Worlali Azumah}
\IEEEauthorblockA{\textit{School of Information Technology} \\
\textit{University of Cincinnati}\\
Cincinnati, Ohio, USA\\
azumahsw@mail.uc.edu}
\and
\IEEEauthorblockN{Nelly Elsayed}
\IEEEauthorblockA{\textit{School of Information Technology} \\
\textit{University of Cincinnati}\\
Cincinnati, Ohio, USA \\
elsayeny@ucmail.uc.edu}
\and
\IEEEauthorblockN{Zag ElSayed}
\IEEEauthorblockA{\textit{School of Information Technology} \\
\textit{University of Cincinnati}\\
Cincinnati, Ohio, USA \\
elsayezs@ucmail.uc.edu}
\and
\IEEEauthorblockN{Murat Ozer}
\IEEEauthorblockA{\textit{School of Information Technology} \\
\textit{University of Cincinnati}\\
Cincinnati, Ohio, USA \\
ozermm@ucmail.uc.edu}
\and
\IEEEauthorblockN{Amanda La Guardia}
\IEEEauthorblockA{\textit{School of Human Services} \\
\textit{University of Cincinnati}\\
Cincinnati, Ohio, USA \\
laguaraa@ucmail.uc.edu}
}

\maketitle

\begin{abstract}
Cyberbullying is a significant concern intricately linked to technology that can find resolution through technological means. Despite its prevalence, technology also provides solutions to mitigate cyberbullying. To address growing concerns regarding the adverse impact of cyberbullying on individuals' online experiences, various online platforms and researchers are actively adopting measures to enhance the safety of digital environments. While researchers persist in crafting detection models to counteract or minimize cyberbullying, malicious actors are deploying adversarial techniques to circumvent these detection methods. This paper focuses on detecting cyberbullying in adversarial attack content within social networking site text data, specifically emphasizing hate speech. Utilizing a deep learning-based approach with a correction algorithm, this paper yielded significant results. An LSTM model with a fixed epoch of 100 demonstrated remarkable performance, achieving high accuracy, precision, recall, F1-score, and AUC-ROC scores of 87.57\%, 88.73\%, 87.57\%, 88.15\%, and 91\% respectively. Additionally, the LSTM model's performance surpassed that of previous studies.

\end{abstract}


\section{Introduction}
Social media platforms allow individuals to stay informed about trends and news and freely express their opinions, fostering user discussions. However, the absence of effective moderation on these platforms has led to various issues, including the rampant dissemination of false information, online harassment, and cyberbullying~\cite{hosseini2017deceiving}. Cyberbullying is a prevalent issue across various online platforms, particularly among young users who share their thoughts, interests, and challenges. Unfortunately, some individuals resort to harmful tactics like harassment, threats, intimidation, and mocking to purposefully inflict emotional harm. The intention behind such behavior is to make others feel worse, erode their self-esteem, and discourage them from engaging in online discussions, posting questions, messages, or sharing personal images~\cite{ptaszynski2018cyberbullying}. Anonymity presents a significant appeal to cyberbullies, as it shields them from social repercussions and negative consequences associated with their improper behavior. This problem has been on the rise since the advent of social networks. A notable incident occurred with the closure of Formspring, likely due to the suicides linked to cyberbullying messages exchanged on that platform~\cite{ptaszynski2018cyberbullying}. The anonymity feature of the service, enabling individuals who know each other offline to message each other anonymously, played a crucial role.

Cyberbullying is a pressing issue deeply rooted in technology. However, technology can also serve as a solution to reduce or even eliminate it~\cite{farley2021cyberbullying}. In response to the increasing worry surrounding the negative effects of cyberbullying on individuals' Internet experiences, numerous online platforms and researchers are implementing measures to improve the safety of their online environments~\cite{Meta}~\cite{Etim_2016}. These online platforms utilize various strategies to address the issue, including refining content through crowdsourcing utilizing upvotes and downvotes, disabling comments, or implementing manual moderation to minimize the impact of inappropriate content~\cite{Greenberg_2017}. However, these approaches have proven to be inefficient and lacking scalability. Consequently, researchers have been significantly demanding to devise methods for automatically detecting abusive or toxic content in real-time~\cite{wulczyn2017ex}. Developing Natural Language Processing (NLP) algorithms specifically designed to detect cyberbullying is one approach towards this goal. Such algorithms rely on annotated data to measure their performance effectively. Popular Machine Learning (ML) algorithms, particularly Deep Neural Networks (DNN), require large annotated corpora to attain high-quality classification results. While researchers continue developing detection models to curb or reduce cyberbullying, attackers use adversarial techniques to bypass detection methods. Machine learning models are typically optimized to perform well when operating with clean data and in benign conditions~\cite{barreno2006can}. However, according to most research works highlighted by Hossein et al.~\cite{hosseini2017deceiving}, they are susceptible to attacks in adversarial techniques. These attacks involve exploiting vulnerabilities in the machine learning algorithms, wherein an adversary can manipulate the algorithm's prediction scores by making slight, often imperceptible perturbations to the input data~\cite{barreno2010security}~\cite{huang2011proceedings}. Such inputs are known as adversarial examples. These inputs are commonly referred to as adversarial examples~\cite{szegedy2013intriguing}, and they have demonstrated their effectiveness in bypassing various machine learning algorithms, even when the adversary has limited knowledge and can only access the target model as a black box~\cite{papernot2017practical}.
This paper aims to improve the user experience when using different social media platforms by investigating and implementing cyberbullying content detection in adversarial text data, utilizing a simple deep learning approach that can be efficiently implemented on different platforms.

\subsection{Adversarial Attacks}
Adversarial attacks involve creating examples with the intent to deceive a system or model. According to Goodfellow et al., ~\cite{goodfellow2014explaining}, adversarial examples are inputs formed by applying small but intentionally worst-case perturbations to examples from the dataset, aiming to make the model output an incorrect answer with high confidence. These attacks can involve subtle or significant modifications to previously correctly classified examples of various types to cause the model to misclassify them while maintaining readability for human users~\cite{goodfellow2014explaining}.
 In the field of Natural Language Processing (NLP), various types of adversarial attacks exist, categorized by Alsmadi et al.~\cite{alsmadi2021adversarial} into four groups:

i) Character level attacks: Involving substitutions, additions, or deletions of characters within a word (typos, adding white spaces, etc.). Example: I want to k ! l l you.
ii) Word-level attacks: Entailing the replacement of words with adversarial synonyms that are more challenging to classify. Example: I want to end you.
iii) Sentence level attacks: Encompassing additions, deletions, or paraphrasing of entire sentences. Example: I would love to send you to heaven today.
iv) Inter-level attacks: A combination of character, word, and sentence-level attacks.
Semantic changes, particularly in word and sentence-level attacks, pose complexity as they require an understanding of grammar and sentence structure. In many instances, human users might not discern these changes as attempts to evade detection~\cite{bitton2022adversarial}. On the other hand, character-level attacks are more straightforward, making alterations only to individual letters.

Adversarial attacks exploit models' weaknesses, and knowledge of the model architecture, functions, and parameters provides an advantage~\cite{alsmadi2021adversarial}. In practical scenarios, however, attackers often lack insight into the specific model they aim to deceive. Consequently, the discovery and exploitation of model-specific weaknesses usually occur through trial and error.
Adversarial attacks can be initiated with various motives, such as influencing a classifier's decision or compromising security. An illustrative case involves the exploitation of vulnerabilities in Microsoft's Tay chatbot, resulting in its shutdown due to the generation of racist tweets~\cite{alsmadi2022adversarial}.

It is worth noting that while each system may have individual weak points, many systems share vulnerabilities to the same types of attacks, as observed by Goodfellow et al.~\cite{goodfellow2014explaining}. These weaknesses can be consistent across various models and training datasets.
\section{Related Work}
Deep learning falls under the umbrella of machine learning and is characterized by computational models with multiple layers, offering a high level of abstraction. These models learn from experience, perceiving the world through a hierarchy of concepts. Utilizing the backpropagation algorithm, deep learning explores intricate details in extensive datasets to compute data representation in each layer based on the preceding layer's representation~\cite{lecun2015deep}.

Its significance lies in providing solutions to previously insurmountable problems with conventional machine learning techniques. Advancements in deep neural network models, coupled with high performance hardware, have driven progress in traditional areas like image classification, speech recognition, and language translation, as well as more sophisticated domains such as drug molecule analysis~\cite{ma2015deep}, brain circuit reconstruction~\cite{helmstaedter2013connectomic}, particle accelerator data analysis~\cite{ciodaro2012online}, and studying DNA mutation effects~\cite{xiong2015human}.

The unprecedented accuracy of deep learning networks has revolutionized AI-based services on the Internet. Major players like Google, Alibaba, Intel, Amazon, and Nvidia have leveraged deep learning for cloud computing AI-based services. Its applications extend to safety and security critical environments, including self-driving cars, malware detection, drones, and robotics. Recent advancements in face recognition have led to biometric authentication in ATMs and mobile phones. Products like Apple Siri, Amazon Alexa, and Microsoft Cortana have become possible through Automatic Speech Recognition (ASR) models and Voice Controllable Systems (VCS).

However, as deep neural networks transition from labs to the real world, concerns arise regarding the security and integrity of applications. Adversaries can manipulate legitimate inputs, which are imperceptible to humans but capable of forcing a trained model to produce incorrect outputs. Szegedy et al.~\cite{szegedy2013intriguing} first identified the susceptibility of well performing deep neural networks to adversarial attacks. Carlini et al.~\cite{carlini2016hidden} and Zhang et al.~\cite{zhang2017dolphinattack} highlighted vulnerabilities in automatic speech recognition and voice controllable systems. Kurakin et al.~\cite{kurakin2016adversarial} demonstrated attacks on autonomous vehicles by manipulating traffic signs. Several countermeasures have been proposed to address adversarial attacks, including adversarial training, distillation, and Generative Adversarial Networks (GANs). However, no single solution has proven effective against all types of attacks, and implementing these defenses may lead to performance degradation and reduced model efficiency.

Hate speech, characterized by abusive or threatening expressions directed against a particular group based on attributes like color, race, religious beliefs, gender, or sexual orientation, is recognized as a significant contributor to escalating global violence~\cite{laub2019hate}. The prevalence of social networks such as Facebook, Instagram, and Twitter, coupled with increased Internet accessibility, has exacerbated the issue by enabling individuals to freely and effectively express their opinions to a global audience~\cite{matamoros2021racism}.

In response to growing concerns about global violence, various initiatives have been undertaken to identify potential sources and mitigate the spread of hate speech on social networks. The AI, ML, data science, and NLP communities have contributed by proposing innovative hate speech detection techniques~\cite{schmidt2017survey}. For example, Mozafari et al.~\cite{mozafari2020bert} introduced a hate speech detection and classification framework for Twitter text streams based on BERT. However, like other NLP applications, hate speech detection methods are susceptible to adversarial attacks, as demonstrated in a study where state-of-the-art NLP adversarial attacks modified text to deceive hate speech recognition models~\cite{oak2019poster}. The objective of these attacks is to disrupt the classification capabilities of the models, resulting in the misclassification of abusive and toxic content.

The literature provides various examples of adversarial attacks on hate speech detection techniques. Gröndahl et al.~\cite{grondahl2018all} employed three types of adversarial attacks: word changes, word-boundary changes, and appending unrelated innocuous words. Similarly, in another study, hate speech detection models were fooled using typos, removing white spaces, inserting benign words, and appending character boundaries~\cite{moh2020no}. Combining individual attack types resulted in more effective adversarial attacks.

To counter attacks on hate speech detection models, several defense strategies have been introduced. Many proposed methods rely on adversarial training to handle perturbations~\cite{grondahl2018all}~\cite{tran2020habertor}~\cite{xia2020demoting}. For instance, adversarial training in~\cite{tran2020habertor} extends the fast gradient method (FGM) attack by incorporating a learnable and fine-grained noise magnitude, adding noise to misleading samples. Additionally, some solutions utilize preprocessing to defend against adversarial attacks on hate speech detection models~\cite{khieu2019tsar}. Moh et al.~\cite{moh2020no} introduced four preprocessing defense techniques for space removal, typos, benign word insertion, and character boundary appending attacks.

Similar to other social media applications, a significant portion of hate speech detection solutions rely on graph-based approaches for robust detection. Beatty et al.~\cite{beatty2020graph} demonstrated the robustness of graph-based models against adversarial attacks, surpassing the performance of text-based solutions in detecting hate or toxic speech~\cite{beatty2020graph}. In~\cite{li2021neighbours}, a graph-based solution was proposed to protect against adversarial attacks by incorporating the concept of latent neighborhood and systematic sampling of neighborhood nodes.

The research of Hosseini et al.~\cite{hosseini2017deceiving} proposes and presents an attack on the Perspective toxic detection system using adversarial examples. The Perspective  API detection system is a recent project developed by  Google and Jigsaw that utilizes machine learning to automatically identify online insults, harassment, and abusive speech~\cite{Perspective_2023}. In their paper, they demonstrated that an adversary can make subtle modifications to highly toxic words, causing the system to assign a significantly lower toxicity score to it. By applying this attack to the sample words provided on the Perspective website, the researchers consistently reduced the toxicity scores to match those of non-toxic words. By conducting various experiments, they demonstrated that an adversary can manipulate the system by intentionally misspelling abusive words or inserting punctuation marks between the letters.

After experimentation, the researchers noticed that the adversarial perturbations can transfer across different words. This means that if a specific modification to a word decreases the toxicity score of one word, applying the same modification to the word is likely to reduce the toxicity score of another phrase as well. Exploiting this characteristic, an adversary can create a dictionary of adversarial perturbations for each word, greatly streamlining the attack process~\cite{hosseini2017deceiving}. Furthermore, it was observed that the Perspective detection system occasionally assigns high toxicity scores to seemingly harmless words incorrectly, assigned a 34\% toxicity score to a majority of misspelled and randomly generated words, and the Perspective interface enables users to provide feedback on the toxicity scores of words, indicating that the learning algorithm incorporates new data to update itself. However, this functionality also opens the system to potential poisoning cyber attacks. In such attacks, an adversary can manipulate the training data, specifically the labels, to trick the model into assigning low toxicity scores to specific words.

These tactics enable the adversary to deceive the Perspective toxic detection system. Such adversarial examples pose a significant threat to toxic detection systems and or cyberbullying detection systems, severely undermining their effectiveness and usability.
Similar to~\cite{hosseini2017deceiving}, Li, Jinfeng, et al.~\cite{li2018textbugger} discusses the security vulnerabilities of Deep Learning based Text Understanding (DLTU) and proposes a general attack framework called TEXTBUGGER for generating adversarial texts. The problem is that DLTU is vulnerable to adversarial text attacks, where maliciously crafted texts trigger target DLTU systems and services to misbehave~\cite{carlini2017towards}. This is concerning given the increasing use of DLTU in security-sensitive applications such as sentiment analysis and toxic content detection. The proposed solution, TEXTBUGGER, is a general attack framework that outperforms state-of-the-art attacks in terms of attack success rate, preserves the utility of benign text, and generates adversarial text with computational complexity sub-linear to the text length. In this paper the authors empirically evaluated TEXTBUGGER on a set of real world DLTU systems and services used for sentiment analysis and toxic content detection, demonstrating its effectiveness, evasiveness, and efficiency. For example, TEXTBUGGER achieves a 100\% success rate on the IMDB dataset using Amazon AWS Comprehend in just 4.61 seconds while still maintaining a 97\% semantic similarity~\cite{li2018textbugger}.

The authors concluded in their paper that TEXTBUGGER is effective and efficient for generating targeted adversarial NLP. The transferability of such examples hint at potential vulnerabilities in many real applications, including text filtering systems and online recommendation systems. The authors suggested possible defense mechanisms to mitigate such attacks, such as spelling checks and adversarial training, and propose exploring an ensemble of linguistically aware or structurally aware-based defense systems to improve robustness.

In one of the initial attempts to deceive deep neural text classifiers, Papernot et al.~\cite{papernot2016crafting} introduced a white box adversarial attack. They iteratively applied the attack to modify an input text until the resulting sequence was misclassified. Although successful in fooling the classifier, their approach resulted in significant alterations at the word level, significantly impacting the original meaning.

In another study by Ebrahimi et al.~\cite{ebrahimi2017hotflip}, they proposed a gradient-based optimization method. This technique involved changing one token to another using the gradients of the model with respect to the one-hot vector input. Additionally, Samanta et al.~\cite{samanta2017towards} utilized embedding gradients to identify crucial words and devised heuristic rules combined with hand-crafted synonyms and typos. These various approaches aimed to manipulate the text inputs to evade classification models while preserving some semantic relevance.

Other prior works also focused on generating adversarial examples for text by substituting a word with another legible but out of vocabulary word~\cite{hosseini2017deceiving}~\cite{belinkov2017synthetic}~\cite{gao2018black}. For example, Belinkov et al.~\cite{belinkov2017synthetic} demonstrated that character-level machine translation systems exhibit heightened sensitivity to random character manipulations, such as keyboard typos. Similarly, Gao et al.~\cite{gao2018black} introduced DeepWordBug, which utilizes character perturbations to create adversarial texts specifically targeting deep learning classifiers. However, it is worth noting that DeepWordBug is not computationally efficient and is not suitable for practical applications~\cite{li2018textbugger}.

Thomas et al.~\cite{hartvigsen2022toxigen} introduced TOXIGEN, a novel dataset comprising 274k toxic and benign statements related to 13 minority groups. This dataset was created to address issues where toxic language detection systems tend to incorrectly flag text as toxic. To achieve this, they devised a demonstration-based prompting framework and employed an adversarial classifier in the loop decoding method to generate subtly toxic and benign text using a large pre-trained language model. By controlling machine generation in this manner, TOXIGEN extends its coverage to implicitly toxic text on a larger scale and across more demographic groups than previous resources relying on human written text. Additionally, they conducted a human evaluation on a challenging subset of TOXIGEN and observed that annotators face difficulty distinguishing machine generated text from human written language. Additionally, our findings indicate that human annotators label 94.5\% of toxic examples as hate speech. Leveraging three publicly available datasets, we demonstrate that fine-tuning a toxicity classifier on our data significantly enhances its performance on human written data. Moreover, we showcase that TOXIGEN proves effective in combating machine-generated toxicity, as fine-tuning substantially improves the classifier's performance on our evaluation subset.

Another study by Grolman et al.~\cite{grolman2022hateversarial} introduced an innovative algorithm designed to generate adversarial examples targeting hate speech detection models in both white box and black box scenarios. The algorithm was specifically tailored to address the unique characteristics of tabular datasets, including immutable features and diverse feature types. By combining these enhancements, this approach achieved significant improvements. In white box attacks, the enhanced method attained nearly 90\% transferability, a substantial enhancement over the baseline attack's 52\% transferability. Similarly, the black box attacks approach achieved 75\% transferability, outperforming the baseline attack's result of 60\%.

Grolman et al.~\cite{grolman2022hateversarial} highlighted a departure from traditional adversarial attack strategies, which primarily rely on altering features based on the substitute model's gradient. Instead, they demonstrated that employing alternative machine learning methods, such as feature importance and correlation-based techniques, yields superior results. Furthermore, they illustrated that successful adversarial attacks can be achieved by modifying user behavior patterns rather than altering the content of their tweets.

Continual challenges persist in hate speech detection due to adversarial attacks, which involve modifying correctly classified examples to induce misclassification, often with the aim of evading detection. These alterations intend to be incomprehensible to a model while remaining easily readable for humans. Another study assesses the robustness of two German hate speech detection models against six distinct character-level adversarial attacks, following the methodology outlined by Gröndahl et al.~\cite{grondahl2018all} in their All You Need Is Love research paper. Despite advancements in models since Gröndahl et al.'s~\cite{grondahl2018all} experiments, even simplistic attacks strongly impact the models. This research aimed to evaluate German hate speech detection systems against character-level adversarial attacks and identify their vulnerabilities. An expectation existed for substantial improvements in hate speech detection models since Gröndahl et al.'s~\cite{grondahl2018all} experiments. However, while improvements have occurred, the models still exhibit susceptibility, particularly to straightforward attacks.

\section{Methodology}
\begin{figure}[htbp]
    \centering
    \includegraphics[width=1.0\linewidth]{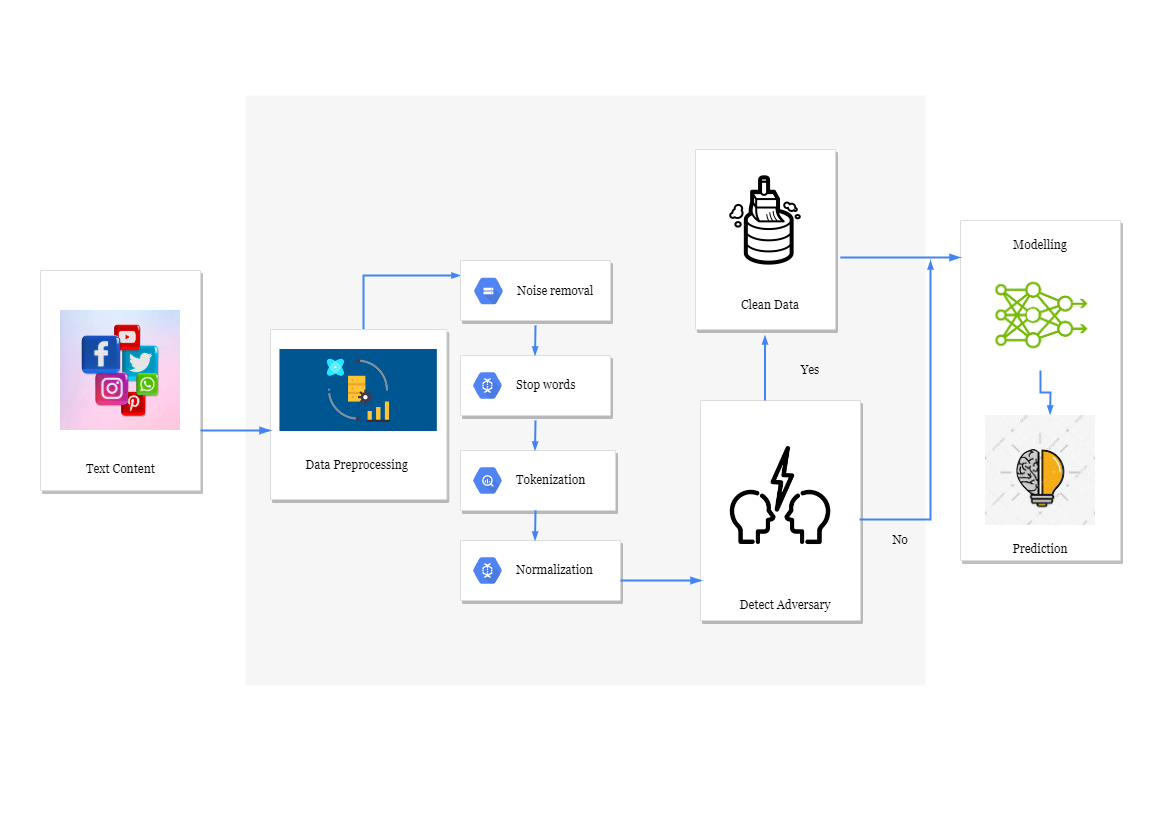}
    \caption{The proposed adversarial and cyberbullying detection methodology in online social networking sites (SNS).}
    \label{fig:enter-label3}
\end{figure}

This section is a proposed approach that outlines the different stages involved in the cyberbullying detection process. Figure~\ref{fig:enter-label3} shows a proposed adversarial and cyberbullying or hate speech detection methodology in online social networking sites (SNS).

\subsection{Dataset} \label{data}
This collection of data, named hate speech offensive, developed by Davidson et al.~\cite{hateoffensive}, has been meticulously assembled and comprises annotated tweets curated explicitly for the purpose of identifying hate speech and offensive language. The data gathering for this dataset involved crowdsourcing by utilizing Twitter's public API to extract tweets using predefined search terms associated with hate speech and offensive language. Subsequently, these tweets underwent manual labeling by multiple annotators who examined them to assign appropriate classifications. The dataset predominantly contains English tweets and serves as a resource for training machine learning models or algorithms to detect hate speech.

The dataset provides various columns containing valuable information for comprehending the classification of each tweet. The column named count represents the total number of annotations for each tweet, while the hate speech count indicates how many annotations classified a particular tweet as hate speech. Conversely, offensive language count denotes the number of annotations categorizing a tweet as containing offensive language. Moreover, neither count reveals how many annotations identified a tweet as neither hate speech nor offensive language.

\subsection{Data Preprocessing}
The first phase of this research entails the creation of an adversarial attack model meticulously designed to identify and correct adversarial examples or attacks within publicly available cyberbullying datasets. This is a critical step in the preprocessing phase of the cyberbullying detection system, aimed at enhancing classification accuracy. This endeavor draws inspiration from existing research on adversarial attacks and aims to replicate the methodologies advanced by pioneering researchers in this domain. To start the first phase, the dataset undergoes several data processing techniques, including normalization, noise removal, stop words removal and tokenization. These preprocessing methods are discussed extensively in the subsections.
Numerous research efforts have emerged to identify adversarial attacks in cyberbullying or hate speech datasets, employing natural language processing (NLP) and machine learning (ML) methods like permutation recovery and obfuscation. This dissertation employs a Spell checker to rectify adversarial input within the dataset before conducting classification, given the textual nature of the data. Spellcheckers are essential components of word processors, capable of identifying and highlighting incorrect words in text. They offer the functionality to correct these errors by suggesting suitable alternatives from a predefined list of words~\cite{singh2018review}. Typically, spellcheckers examine each word in a text individually, comparing it against a stored lexicon to determine its correctness. If a word is found in the lexicon, it is deemed appropriate regardless of its context~\cite{singh2018review}. Additionally, several studies have explored the effectiveness of spelling and grammatical checks in defending against character or word level attacks in text data, which involve modifying characters and words~\cite{alsmadi2022adversarial}. 
To assess the resilience of our adversary detection and correction phase, the spell checker was incorporated into the data preprocessing stage. To be specific, Python's SpellChecker library was employed for this purpose.

\subsubsection{Noise Removal}
Noise removal removes URL’s multiple spaces, punctuation, hashtags, usernames, and emojis from the text data for processing. This method is essential because it helps to reduce noise in the dataset and focuses mainly on the words that carry the most meanings. 
\subsubsection{Stop Words}
Stop words are often used as function words, such as articles, conjunctions, and prepositions, and can make text higher dimensional. Removing stop words can reduce the dimensionality of term space, making it easier to analyze and interpret text data. It can also improve the accuracy of NLP models. This is because stop words can often be misleading to models, and they can cause the models to make incorrect predictions. 
\subsubsection{Tokenization}
Tokenization breaks down text into smaller units called tokens by splitting the text into paragraphs, sentences, and words. Tokenization is used because it is impossible to feed the whole text sample to a model at once. By breaking the text down into smaller units, the model can process the text more easily~\cite{kumar2022study}. Additionally, the meaning of the text is preserved when it is tokenized, so the model can still understand the text. This method is essential because many NLP algorithms rely on individual words as the basis of analysis, so breaking down text into its individual components is necessary.
\subsubsection{Normalization}
Normalization is converting all the tokens in a text to a standard form. This can involve lower casing all the letters, removing stop words, and stemming or lemmatizing the words. These steps are undertaken to preprocess the data for both the adversary and cyberbullying detection model. The methodology is divided into two key aspects: the adversarial example detection phase, designed to identify and rectify adversarial examples, and the second phase, which introduces a cyberbullying detection model utilizing the corrected data from the initial phase.
\subsection{Deep Learning Model Architecture}
The second phase introduces an innovative cyberbullying detection model, leveraging deep learning techniques specifically tailored for text detection. The model will harness publicly available data sourced from social networking sites, as mentioned in the first phase. This phase represents a forward-looking stride in the ongoing pursuit of advancing cyberbullying detection, striving to enhance the effectiveness and accuracy of identifying and mitigating this pervasive online issue.
\subsubsection{Long Short Term Memory (LSTM)}
The Long Short-Term Memory (LSTM) is a category of recurrent neural network (RNN) designed for classifying and predicting temporal dependent data, such as time series and signal datasets~\cite{roopak2019deep}~\cite{elsayed2019gated}~\cite{azumah2021deep}. Initially applied in natural language processing (NLP) and speech recognition, LSTM has demonstrated superior learning precision when compared to other algorithms like recurrent cascade correlation and neural sequence chunking~\cite{sak2014long}~\cite{sutskever2014sequence}.

Research findings indicate that LSTM effectively handles tasks that were challenging for traditional recurrent network algorithms in the past~\cite{hochreiter1997long,elsayed2022litelstm}. Introduced by Hochreiter et al.~\cite{fischer2018deep}, the LSTM architecture comprises the input gate, a memory cell, and an output gate~\cite{hochreiter1997long}. Within an LSTM network, memory cells store features and consist of a forget gate ($f_t$), input gate ($i_t$), and an output gate ($o_t$)~\cite{hochreiter1997long}.
\begin{figure}[htbp]
    \centering
    \includegraphics[width=1\linewidth]{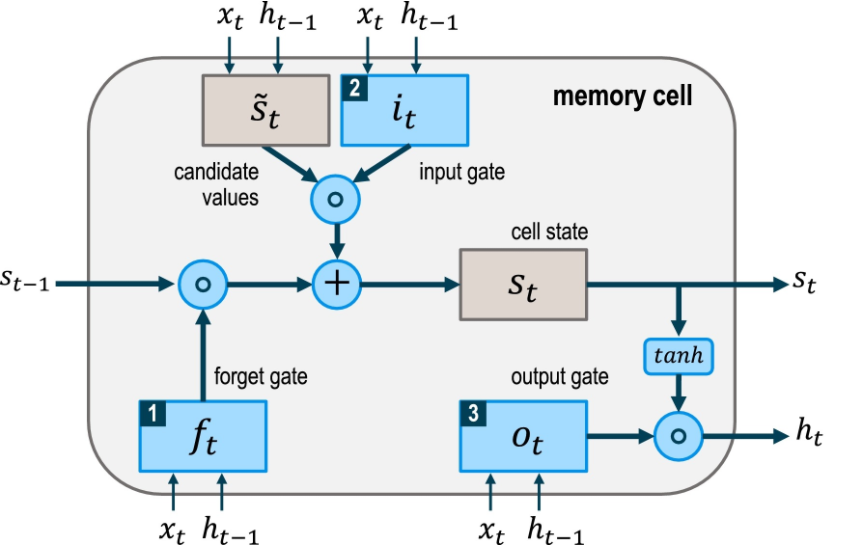}
    \caption{Long Short Term Memory (LSTM) block architecture or model for classification of cyberbullying text in online social networking sites (SNS)~\cite{azumah2021deep}.}
    \label{fig:enter-label7}
\end{figure}

Figure~\ref{fig:enter-label7} illustrates the design of the Long short term memory cell's architecture as follows:
\begin{itemize}
    \item \textit{Forget gate $(f)$}: Specifies the details regarding the removal of information from the cell state~\cite{azumah2021deep}.
    \item \textit{Input gate $(i)$}: Determines the inclusion of information into the cell state~\cite{azumah2021deep}.
    \item \textit{Input update $(o)$}: Modifies the memory content according to the present input to the LSTM~\cite{azumah2021deep}.  
    \item \textit{Output gate $(o)$}: Establishes and outlines the information to be presented as output from the cell state~\cite{azumah2021deep}.
\end{itemize}
At each time-step $t$, the computation of each component in the standard LSTM is determined as follows:
\begin{align}
i_{(t)}&= \sigma(W_{xi} x_{(t)} +U_{hi}  h_{(t-1)}+ b_i)\label{eqn:i_lstm}\\ 
g_{(t)}&= \mathrm{tanh}(W_{xg} x_{(t)} +U_{hg}  h_{(t-1)}+ b_g)\label{eqn:g_lstm}\\ 
f_{(t)}&=\sigma(W_{xf} x_{(t)} +U_{hf}  h_{(t-1)} + b_f)\label{eqn:f_lstm}\\ 
o_{(t)}&=\sigma(W_{xo} x_{(t)} +U_{ho}  h_{(t-1)} + b_o)\label{eqn:o_lstm}\\ 
s_{(t)}&= f^{(t)}\odot s_{(t-1)} + i_{(t)} \odot g_{(t)}\label{eqn:s_lstm}\\
h_{(t)}&= \mathrm{tanh}(s_{(t)})\odot g_{(t)}\label{eqn:h_lstm}
\end{align}
\noindent
where, $x_{(t)}$ represents the input at time step $t$, and $h_{(t-1)}$ denotes the output of the memory cells from the previous time-step $t-1$~\cite{fischer2018deep}. The symbols $\sigma$ and $\odot$ correspond to the logistic sigmoid function and element-wise Hadamard multiplication, respectively. The model incorporates two activation units: input-update and output activation, where the $\mathrm{tanh}$ activation function is recommended for use~\cite{elsayed2018a}.

The memory cell state at time $t$ is denoted as $s^{(t)}$, and the output of the LSTM unit at time $t$ is $h_{(t)}$. The biases for each gate are represented by $b_i$, $b_g$, $b_f$, and $b_q$. The feed forward weights and recurrent weights are denoted by $W$ and $U$, respectively.

\section{Experiment, Results And Analysis}
\subsection{Training And Testing}
In this study, the Long Short Term Memory (LSTM) model uses a 60-20-20 split ratio to train, validate and test. The model is trained on 80\% of the dataset, validated on 20\% and tested with 20\% of the dataset.
\subsection{Performance Metrics}
The LSTM model's performance in classifying attacks on cyberbullying data is assessed using standard metrics: Accuracy, Recall, Precision, F1-score, and AUC-ROC. These metrics are calculated as follows:
\begin{align}
Accuracy &=\frac{TP + TN}{TP + TN +FP + FN}\\ 
Recall &=\frac{TP }{TP + FN}\\
Precision &=\frac{TP}{TP + FP}\\
F1 &= 2 \times \frac{Precision * Recall}{Precision + Recall}
\end{align}
\noindent
where TP, TN, FN, and FP represent true positives, true negatives, false negatives, and false positives. Additionally, to check the model's classification performance, we calculate the Area Under the Curve (AUC) - Receiver operating characteristics (ROC) curve.
\subsection{Results}
\begin{figure}[htbp]
    \centering
    \includegraphics[width=1\linewidth]{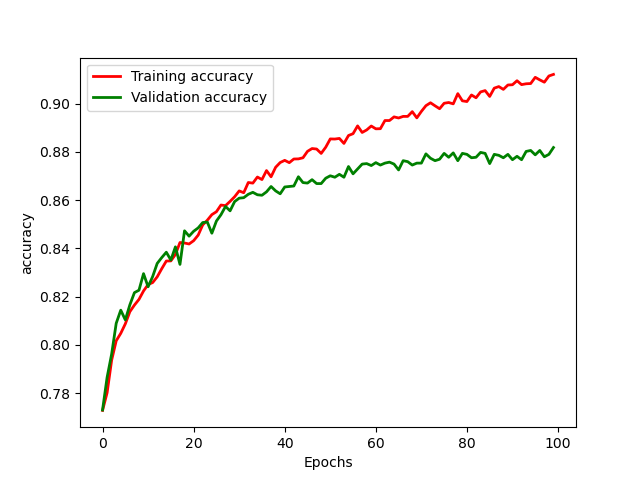}
    \caption{The proposed LSTM detection model training vs. validation accuracy.}
    \label{fig:enter-label8}
\end{figure}

\begin{figure}[htbp]
    \centering
    \includegraphics[width=1\linewidth]{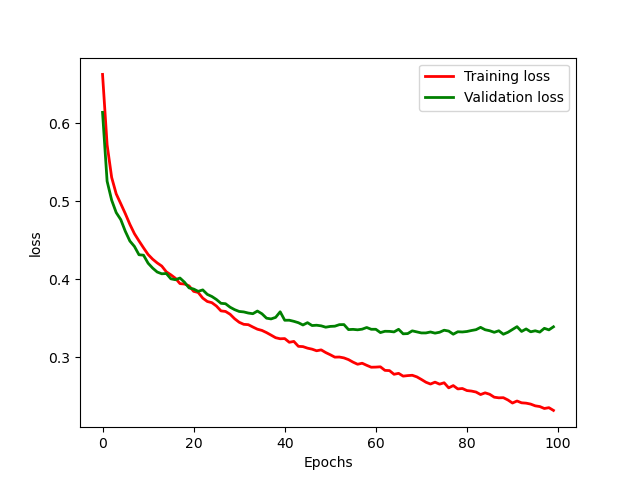}
    \caption{The proposed LSTM detection model training vs. validation loss.}
    \label{fig:enter-label9}
\end{figure}

The outcomes of the experiment are displayed in Tables~\ref{tab:my-table6} and~\ref{tab:my-table7}  where Table~\ref{tab:my-table6} illustrates the experimental results and Table~\ref{tab:my-table7} 
illustrates comparative studies of results. Following the training and testing of the model using the hate speech dataset, the experiment was conducted. The trial commenced and concluded with a consistent epoch of 100, with a progressive augmentation of the model's hidden layers and denseness. Figures~\ref{fig:enter-label8} and~\ref{fig:enter-label9} depict the training and validation accuracy, and the training and validation loss of the model, respectively.

\begin{table}[htbp]
\caption{Our findings  for identifying adversarial attacks and classification occurrences in social networking sites cyberbullying content.}
\begin{center}
\begin{tabular}{|l|c|}
\hline
\textbf{Metrics} & \textbf{Training Results}\\
\hline
Accuracy& 87.57\%\\
Precision&88.73\%\\
Recall&87.57\%\\
F1-Score&88.17\%\\
AUC-ROC&91\%\\
Training parameters&100,227\\
All parameters&2,144,127\\
\hline
\end{tabular}
\label{tab:my-table6}
\end{center}
\end{table}

Table~\ref{tab:my-table6} illustrates the outcomes of the trial experiment conducted using the LSTM model, employing a fixed epoch of 100, yielding Accuracy, Precision, Recall, F1-score, and AUC-ROC scores of 87.57\%, 88.73\%, 87.57\%, 88.17\%, and 91\%, respectively. Model parameters are characteristics of the training data that are learned during the training process. In the context of deep learning, these parameters typically refer to weights and biases. Parameters serve as a gauge of the model's performance. Specifically, in the LSTM model, there are 100,227 trainable parameters and a total of 2,144,127 parameters. The total parameter count represents the sum of all the weights and biases in the LSTM model, while the trainable parameters refer to those that can be learned and adjusted throughout the training process. Meanwhile, the results depicted in Figure~\ref{fig:enter-label8} and Figure~\ref{fig:enter-label9} present the average training and validation loss, along with accuracy. Specifically, the training loss is observed to be 0.0654, with an approximate accuracy of 98\%, while the validation accuracy stands at 88\% with a corresponding validation loss of 0.0652.
\subsection{Comparative Study}
In this section, we conduct a comparative analysis of our research findings in relation to recent studies. Our investigation focuses on the performance of the LSTM model with multiple dense layers in the trial phase. The outcomes reveal significant success, with recorded metrics including an Accuracy of 87.57\%, Precision of 88.73\%, Recall of 87.57\%, F1-Score of 88.17\%, and AUC-ROC of 91\%. To assess the effectiveness of the LSTM model, we compare its results with those of Davidson et al.~\cite{davidson2017automated}, Del et al.~\cite{del2017hate}, and Li et al.~\cite{li2018textbugger}  because these two studies used the same data mentioned in subsection~\ref{data}. Remarkably, our LSTM model surpasses state-of-the-art classification models such as Support Vector Machine (SVM) and Long Short Term Memory (LSTM), as illustrated in Table~\ref{tab:my-table7}.

Davidson et al.~\cite{davidson2017automated} conducted research utilizing the SVM model in conjunction with lexicons for the classification of hate speech, cyberbullying, and offensive language. Their model achieved an Accuracy of 86\%. The researchers further assessed their model's performance using precision, recall, and f1-Score as effective metrics. Despite achieving an overall Precision of 91\%, Recall of 90\%, and F1-score of 90\%, they observed that their model misclassified nearly 40\% of hate speech instances. They concluded that this discrepancy stemmed from the model's bias toward categorizing tweets as less hateful or offensive, leading to a significant under representation of tweets classified as more offensive or hateful than their actual category.

\begin{table}[htbp]
\caption{Our findings were contrasted with the outcomes from recent studies focusing on methods employed for identifying adversarial attacks and classification occurrences in social networking sites cyberbullying content.}
\begin{center}
\begin{tabular}{|l|c|c|}
\hline
\textbf{Model} & \textbf{Method}& \textbf{Accuracy (\%)} \\
\hline
Davidson et al.~\cite{davidson2017automated}&SVM& 86\%\\
Del et al.~\cite{del2017hate}&SVM&75.23\%\\
Our Model &LSTM& 87.57\%\\
\hline
\end{tabular}
\label{tab:my-table7}
\end{center}
\end{table}

Del et al.~\cite{del2017hate} conducted experiments on textual data aimed at classifying hate speech, resulting in an Accuracy of 75.23\% when utilizing both SVM and LSTM models. However, neither SVM nor LSTM could effectively distinguish between the three classes (strong hate, weak hate, and no hate), especially struggling with the classification of strong hate instances. The researchers suggested that this difficulty might stem from the limited number of strong hate documents, which constituted the class with the fewest documents, as well as the low level of agreement among annotators. Consequently, the authors proceeded to conduct an additional experiment focusing on a two-class classification, yielding significantly improved accuracy results compared to their initial trial.

When comparing our model to those employed in the aforementioned studies for hate speech classification, it became apparent that none of the prior works integrated an adversary correction model to rectify inputs or examples that might lead the model to misclassify outputs. In contrast, our study not only assessed the model's performance based on accuracy, precision, recall, and f1-score but also included an evaluation of AUC-ROC. AUC-ROC is regarded as one of the most crucial evaluation metrics for assessing classification models' performance~\cite{narkhede2018understanding}. It provides a comprehensive measure of overall accuracy during model testing.

The AUC-ROC curve ranges from 0 to 1, where 0 signifies a perfectly inaccurate test and 1 signifies a perfectly accurate test~\cite{mandrekar2010receiver}. Typically, an AUC of 0.5 indicates no discrimination (i.e., the ability to classify correctly), while values between 0.7 and 0.8 are considered acceptable, 0.8 to 0.9 are deemed excellent, and above 0.9 are regarded as outstanding~\cite{mandrekar2010receiver}.

In our study, the AUC-ROC score achieved using the LSTM model is 91\%, indicating a 91\% probability that the model accurately distinguished among the three classes: hate speech, offensive language, and neither. This underscores the effectiveness of our model in tackling hate speech classification.

\begin{figure}[htbp]
    \centering
    \includegraphics[width=1\linewidth]{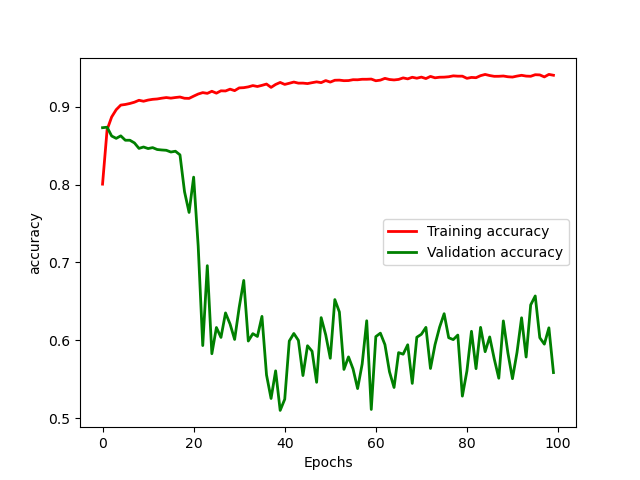}
    \caption{1D Convolutional Neural Network (CNN) detection model training versus validation loss.}
    \label{fig:enter-label10}
\end{figure}

\begin{figure}[htbp]
    \centering
    \includegraphics[width=1\linewidth]{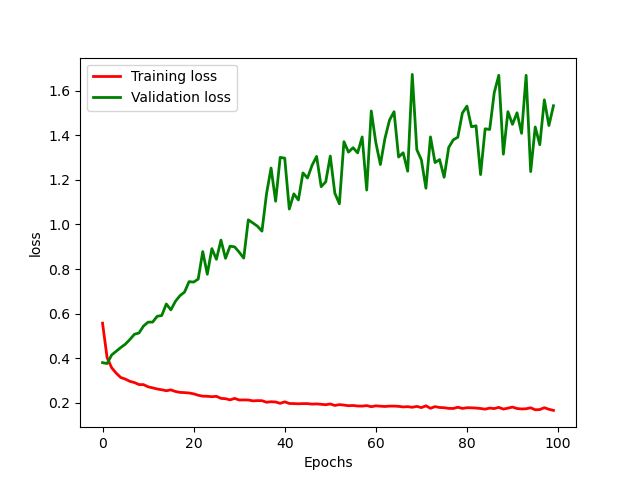}
    \caption{1D Convolutional Neural Network (CNN) detection model training versus validation accuracy.}
    \label{fig:enter-label11}
\end{figure}

\begin{table}[htbp]
\caption{The empirical findings were contrasted with the outcomes from the state-of-the-art deep learning model focusing on methods employed for identifying adversarial attacks and classification occurrences in social networking sites cyberbullying content.}
\begin{center}
\begin{tabular}{|l|c|c|}
\hline
\textbf{Model} & \textbf{Method}& \textbf{Accuracy (\%)} \\
\hline
Model 1& 1D-CNN&55.27\%\\
Model 2& GRU&84.30\%\\
Our Model&LSTM&87.57\%\\
\hline
\end{tabular}
\label{tab:my-table11}
\end{center}
\end{table}

\begin{figure}[htbp]
    \centering
    \includegraphics[width=1\linewidth]{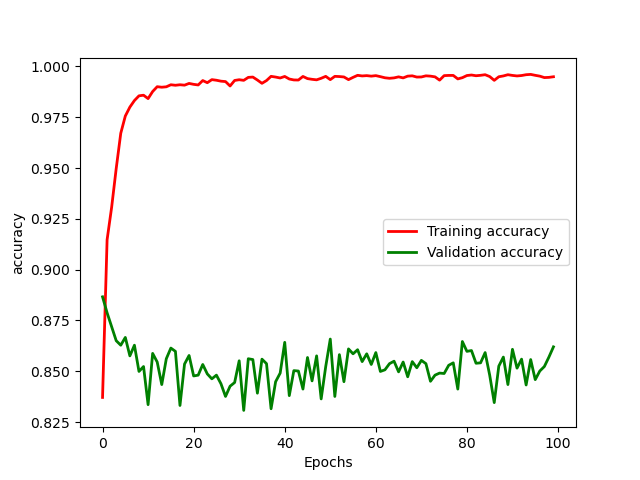}
    \caption{GRU detection model training versus validation accuracy.}
    \label{fig:enter-label12}
\end{figure}

\begin{figure}[htbp]
    \centering
    \includegraphics[width=1\linewidth]{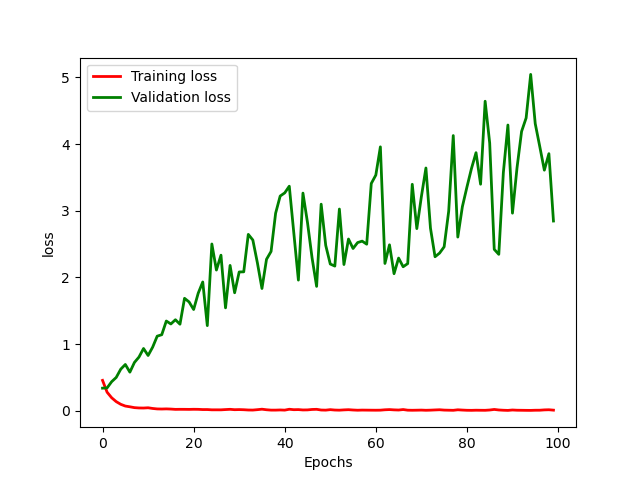}
    \caption{GRU detection model training versus validation loss.}
    \label{fig:enter-label13}
\end{figure}

In this comparative study, we also applied both the GRU and 1D-CNN methods to our dataset. Employing a fixed epoch of 100 and conducting a single trial experiment, we observed the following performance metrics: For the 1D-CNN model, the Accuracy was 55.27\%, Precision score was 55.98\%, Recall was 55.27\%, F1-score was 55.62\%, and AUC-ROC was 66.77\%. Conversely, the GRU model outperformed with an Accuracy of 84.30\%, Precision of 85.81\%, Recall of 84.30\%, F1-score of 85.05\%, and AUC-ROC score of 88.66\%. Overall, the proposed LSTM model performed better than GRU and 1D-CNN. The results are illustrated in Table~\ref{tab:my-table11}. Visual representations of the training versus validation accuracy and loss for both models are provided in Figure~\ref{fig:enter-label10} and Figure~\ref{fig:enter-label11} for the 1D-CNN, and Figure~\ref{fig:enter-label12} and Figure~\ref{fig:enter-label13} for the GRU model.

\section{Conclusion}
This paper focused on detecting adversarial attacks within social networking site text data, with a particular emphasis on identifying hate speech. It involved an experiment employing a deep learning-based approach combined with a correction algorithm aimed at rectifying adversarial inputs or attacks, ultimately leading to the classification of the corrected text. The trial experiment utilized an LSTM model with a fixed epoch of 100, yielding Accuracy, precision, Recall, F1-score, and AUC-ROC scores of 87.57\%, 88.73\%, 87.57\%, 88.17\%, and 91\%, respectively.

Upon comparing our model with those employed in prior studies for hate speech classification, it was evident that none of the previous works integrated an adversary correction model to rectify inputs that might lead to misclassification. In contrast, our study not only assessed the model's performance based on standard metrics like Accuracy, precision, Recall, and F1-score but also included an evaluation of AUC-ROC, a crucial metric for assessing classification models' performance. The achieved AUC-ROC score of 91\% using the LSTM model indicated a 91\% probability of accurately distinguishing among the three classes: hate speech, offensive language, and neither, showcasing the effectiveness of our model in hate speech classification.

\bibliographystyle{ieeetr}
\bibliography{references}

\end{document}